\documentclass[accepted]{melba}

\usepackage{amsmath,amsfonts}

\usepackage{amssymb}
\usepackage{array}
\usepackage{xurl}
\usepackage{graphicx}
\usepackage{subcaption}
\usepackage{tabularray}
\UseTblrLibrary{booktabs}
\usepackage{pdflscape}
\usepackage{color}
\usepackage{float}
\usepackage{ulem}
\usepackage{tabularx}

\definecolor{green1}{RGB}{141,212,160}
\definecolor{blue1}{RGB}{164,193,228}
\definecolor{purple1}{RGB}{207,188,243}
\definecolor{yellow}{RGB}{191,191,61}
\definecolor{purple2}{RGB}{117,20,124}
\definecolor{green2}{RGB}{57,126,37}
\definecolor{blue2}{RGB}{2,2,242}
\definecolor{red}{RGB}{224,50,37}
\definecolor{orange}{RGB}{255,128,0}

\definecolor{quali_yellow}{RGB}{236, 225, 51}
\definecolor{quali_orange}{RGB}{213, 94, 0}
\definecolor{quali_blue}{RGB}{86, 180, 233}

\melbaid{2026:012}  
\doi{10.59275/j.melba.2026-67g5}
\melbaauthors{Sarah {de Boer} et al}  
\email{sarah.deboer@radboudumc.nl}
\volume{2026}
\firstpageno{229}  
\melbayear{2026}  
\datesubmitted{2026-01-30}  
\datepublished{2026-05-22}  

\ShortHeadings{Robust Renal Mass Segmentation}{{de Boer} et al}

\title{Robust Renal Mass Segmentation on CT: A Validation Study of an AI-Based Framework}

\author{
	\firstname Sarah \surname {de Boer} \aff{1}\orcid{0000-0001-5184-4340},
	\firstname Hartmut \surname H{\"a}ntze \aff{1,2}\orcid{0000-0002-6204-571X},
    \firstname Kiran Vaidhya \surname Venkadesh\aff{1}\orcid{0000-0002-4846-9049},
    \firstname Myrthe A. D. \surname Buser \aff{1}\orcid{0000-0003-0640-6434},
    \firstname Gabriel E. \surname Humpire Mamani \aff{1}\orcid{0000-0002-3199-8163},
    \firstname Lina \surname Xu \aff{2},
    \firstname Lisa C. \surname Adams \aff{4}\orcid{0000-0001-5836-4542},
    \firstname Jawed \surname Nawabi \aff{3}\orcid{0000-0002-1137-0643},
    \firstname Keno K. \surname Bressem \aff{4,5}\orcid{0000-0001-9249-8624},
    \firstname Bram \surname {van Ginneken} \aff{1,6}\orcid{0000-0003-2028-8972},
    \firstname Mathias \surname Prokop \aff{1}\orcid{0000-0001-8157-8055},
    \firstname Alessa \surname Hering \aff{1}\orcid{0000-0002-7602-803X}
}
\affiliations{
	\num 1 \addr Department of Medical Imaging, Radboudumc, Nijmegen, The Netherlands \\
	\num 2 \addr Department of Radiology, Charité - Universitätsmedizin Berlin, Berlin, Germany \\
	\num 3 \addr Department of Neuroradiology, Charité - Universitätsmedizin Berlin, Berlin, Germany \\
    \num 4 \addr Department of Diagnostic and Interventional Radiology, Klinikum rechts der Isar,  TUM University Hospital, Technical University of Munich, Munich, Germany \\
    \num 5 \addr Department of Cardiovascular Radiology and Nuclear Medicine, German Heart Center, TUM University Hospital, Technical University of Munich, Munich, Germany \\
    \num 6 \addr Fraunhofer MEVIS, Bremen, Germany
}

\abstract{
	Renal mass segmentation has important potential to enhance the clinical workflow,  especially in settings requiring quantitative assessments. Kidney volume could serve as an important biomarker for renal diseases, with changes in volume correlating directly with kidney function. Currently, clinical practice often relies on subjective visual assessment for evaluating kidney size and kidney lesions, including tumors and cysts, which are typically staged based on diameter, volume, and anatomical location. To support a more objective and reproducible approach, this research aims to develop a robust, thoroughly validated renal mass segmentation algorithm, named Renal-Net. We employ publicly available training datasets and leverage the state-of-the-art medical image segmentation framework nnU-Net. Validation is conducted using both proprietary and public test datasets, with segmentation performance quantified by Dice coefficient and the 95th percentile Hausdorff distance. Furthermore, we analyze robustness across subgroups based on patient sex, age, CT contrast phases, and tumor histologic subtypes. Our findings demonstrate that our segmentation algorithm, trained exclusively on publicly available data, generalizes effectively to external test sets and outperforms existing state-of-the-art models across all tested datasets. Subgroup analyses reveal consistent high performance, indicating strong robustness and reliability. The developed algorithm and associated code are publicly accessible at \url{https://github.com/DIAGNijmegen/oncology-kidney-abnormality-segmentation}.
}


\begin{document}

\twocolumn[\maketitle]

\section{Introduction}
\label{introduction}

    \enluminure{K}idney cancer has a global incidence rate of approximately 400,000 new cases annually, leading to 175,000 deaths \citep{cirillo_2024a}. 
    It is often detected incidentally during imaging performed for unrelated medical reasons, most often in computed tomography (CT). 
    Treatment options for suspected malignant renal masses include radical and partial nephrectomy \citep{motzer_2022}.
    Ablation techniques are also offered in various medical facilities as a less invasive treatment option for small tumors \citep{filippiadis_2019, mershon_2020}.
    For less aggressive tumors and for patients with reduced life expectancy due to other reasons like age or comorbidities, active surveillance is a treatment alternative \citep{capitanio_2016, kutikov_2010}. 
    Given the variety of available treatment options, patient stratification plays a crucial role in optimizing clinical decision-making.
    In clinical practice, the evaluation of the kidneys and lesions relies on the subjective visual assessment of kidney volume, tumor longest diameter and tumor location which are used to accurately stratify patients \citep{kutikov_2009}. 
    Furthermore, in other kidney diseases, like autosomal dominant polycystic kidney disease (ADPKD), the assessment of total kidney volume (TKV) is an indicator for disease severity and disease progression \citep{gaur_2019}.

    Manual segmentation of organs and tumors on CT is time consuming and the accuracy and reproducibility of the segmentations are subject to inter-rater variability \citep{joskowicz_2019, meyer_2006}. This might be caused by differences in experience levels, but also imaging characteristics. 
    In the medical imaging field, many research has been done on AI-based segmentation of organs of interest and tumor regions \citep{bilic_2023, staal_2004, litjens_2014}. 
    Within the last few years researchers have taken immense steps towards solving this task, starting with the U-Net \citep{ronneberger_2015a} and resulting in the nnU-Net \citep{isensee_2021}.
    The nnU-Net is an automated, self-configuring segmentation model that optimizes its hyperparameters based on a given annotated dataset and is trained in a supervised learning setting. Depending on the specific segmentation task, the size and characteristics of available training data, and hardware resources, users can select among multiple configurations provided by the nnU-Net.
    Multi-organ segmentation models built upon nnU-Net have emerged, with TotalSegmentator \citep{wasserthal_2023} being the most widely adopted framework.

    While general medical image segmentation models have made considerable progress, they may not always perform optimally for specialized tasks such as renal mass segmentation. In such contexts, organ-specific models can achieve superior performance through the use of carefully curated training data and the integration of domain-specific knowledge. Kidney tumors often considerably alter the organ’s shape, size, and internal structures, posing challenges for general segmentation models primarily trained on healthy or standard anatomy. Such structural variations necessitate the development of dedicated AI methods tailored to specific clinical scenarios. For instance, studies have specifically addressed segmentation of large kidney tumors \citep{yang_2022}, while other works have targeted segmentation of kidney lesions in patients with ADPKD \citep{rombolotti_2022}. Beyond tumor and lesion segmentation, AI research has explored additional kidney-related segmentation tasks, such as delineating the renal cortex and medulla for assessing kidney donor candidacy \citep{korfiatis_2022}, and segmenting renal veins and arteries to facilitate surgical planning \citep{khan_2025}. Moreover, general tumor segmentation models, which are not specific to the kidneys, have also been investigated \citep{pang_2020, degrauw_2025}. 
    
    Kidney cancer diagnosis not only relies on CT scans, but Magnetic Resonance (MR) imaging is used for, among other reasons, determining tumor subtypes and characterizing kidney cysts \citep{ljungberg_2019}. U-Net based algorithms for kidney tumor and cyst segmentation on MRI have been proposed \citep{gregory_2021, haghighi_2018a, zollner_2021a}, as well as segmentation algorithms for total kidney volume segmentation \citep{raj_2022} and liver cyst segmentation for patients with ADPKD \citep{chookhachizadehmoghadam_2024}. Additionally, kidney tumor models for MRI can be developed more quickly if equivalent CT models already exist \citep{hantze_2025}. Finally, the inclusion of Positron Emission Tomography (PET) imaging can improve kidney segmentation performance on CT \citep{leube_2024}.

    \begin{figure*}[h!]
        \centering
        \includegraphics[width=1\linewidth]{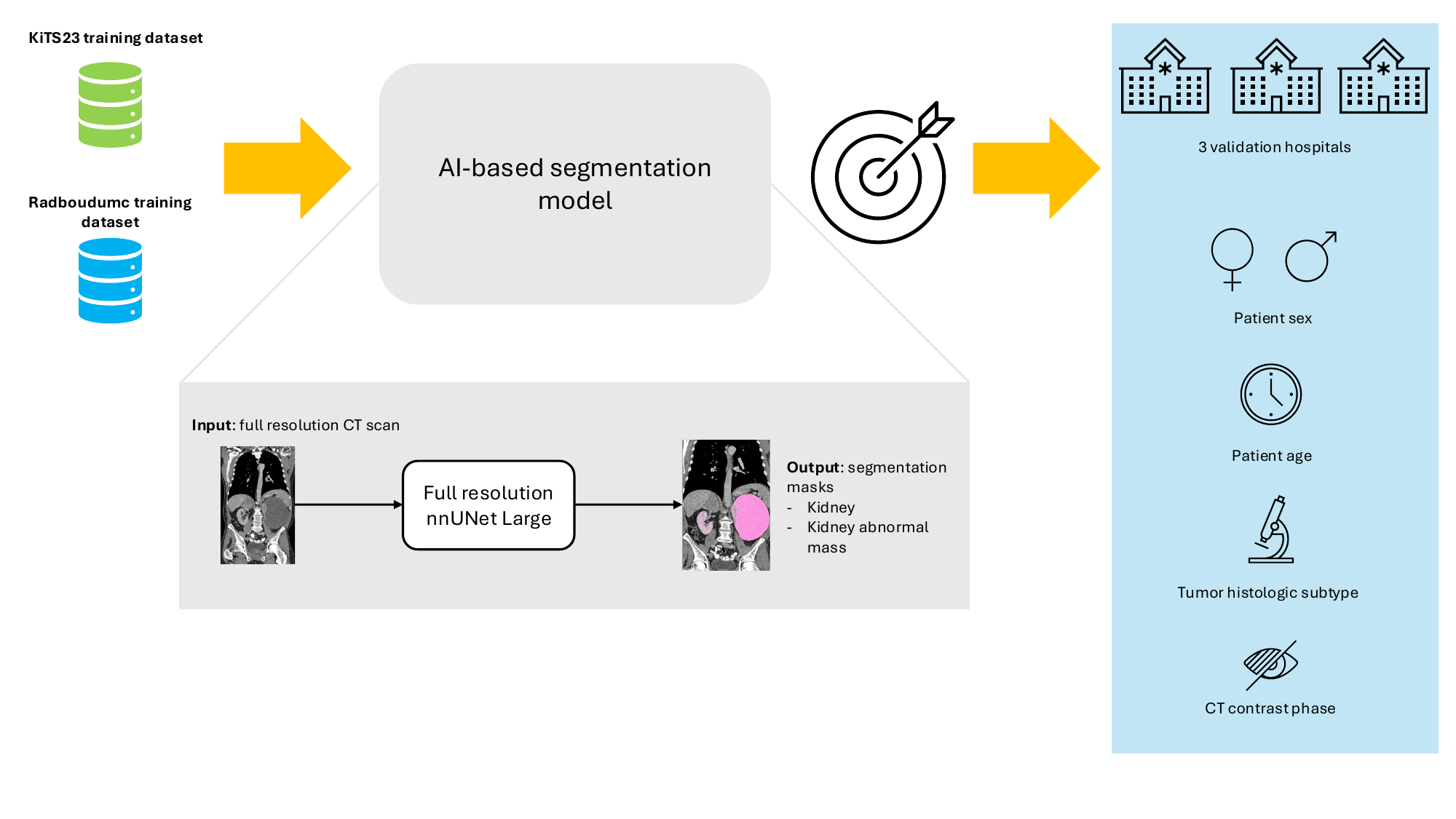}
        \caption{Overview of the research. The proposed AI-based segmentation model is trained on two publicly available datasets. The proposed method consists of a trained nnU-Net that segments the kidney and lesions in the kidney if present. We thoroughly validate the proposed method on three test datasets from three different medical centers, and test performance across patient sex, patient age, tumor histologic subtype and CT contrast phase.}
        \label{figure:full_pipeline}
    \end{figure*}

    The Kidney Tumor Segmentation (KiTS) challenge has considerably advanced the development of segmentation models for kidney cancer \citep{heller_2021, heller_2023}.
    Held three times to date, the challenge has produced numerous studies detailing methods for achieving high performance on the challenge dataset. The KiTS dataset was utilized to train an AI model to generate annotations for various cancer collections from the National Cancer Institute (NCI) Imaging Data Commons (IDC) \citep{murugesan_2024}. Furthermore, an nnU-Net model trained on the KiTS dataset and evaluated on in-house data, and vice versa, demonstrated a decline in external performance \citep{raman_2025}. This performance drop underscores the need for more heterogeneous training datasets and diverse, clinically relevant, external validation cohorts.
    The winners of the 2023 version of the KiTS challenge introduced Auto3DSeg \citep{myronenko_2024}, which is now integrated in the MONAI \citep{consortiummonai_2024} framework along with an implementation for training it on the KiTS dataset. 
    However, recently, nn-UNet is shown to be the framework that leads to state-of-the-art performance on well known benchmarks, while the out-of-the-box performance of Auto3DSeg underperformed compared to nnU-Net \citep{isensee_2024}. On the KiTS23 dataset, nnU-Net ResEnc L achieved a Dice score of 88\%, while Auto3DSeg SegResNet achieved 81\%. The runner up on the KiTS23 challenge investigated different 3D-UNet setups and introduced a multi-scale postprocessing strategy \citep{uhm_2024b}. The third place of the challenge introduced a cascaded approach where an low resolution network first segments the kidneys and an ROI is used as input to the second network making high resolution predictions \citep{george_2022}.

    This research is designed as a large-scale validation study rather than a methodological contribution, addressing a critical gap between algorithmic development and clinical adoption \citep{chouvarda_2025, mercaldo_2025}. Our work makes several distinctive contributions: (1) We develop a robust kidney and renal mass segmentation model (named Renal-Net) that maintains performance across diverse patient populations and imaging protocols; (2) We provide comprehensive validation across three independent datasets totaling over 1,500 CT scans, representing, to the best of our knowledge, the largest external validation in this domain; (3) We conduct detailed subgroup analyses to identify potential biases in model performance across patient demographics, imaging parameters, and tumor characteristics; and (4) We make our validated model publicly available to facilitate clinical testing and further research. The algorithm is available on grand-challenge.org for easy use (\url{https://grand-challenge.org/algorithms/kidney-abnormality-segmentation/}) and our code is available on Github: \url{https://github.com/DIAGNijmegen/oncology-kidney-abnormality-segmentation}.

\section{Methods}\label{methods}
    In this section, the training (Section~\ref{training-datasets}) and testing (Section~\ref{external-validation}) datasets are introduced. We outline how we combined the different training datasets and show the characteristics of all datasets used in this study. Furthermore, we introduce the deep learning framework (Section~\ref{deep-learning-framework}) and layout our post-processing steps. The methodology is outlined in Figure~\ref{figure:full_pipeline}.

    \begin{table*}[!h]
    \centering
    \small
    \caption{Data set characteristics, for the training (KiTS and Radboudumc training set) and testing (Radboudumc test set, TCGA-KIRC, and Charité) datasets. Besides country of origin and number of scans, presented are the mean values averaged over the cases followed by the standard deviations. Lesion and kidney volumes are averaged over the number of lesions and kidneys in that case. Kidney volumes include lesion volumes. Although the Radboudumc dataset contains cases without lesions, these lesion-free cases were excluded when calculating the average lesion count and lesion volumes. However, these cases were included when calculating average kidney volumes, resulting in lower average kidney volumes compared to the other datasets.}
    \begin{booktabs}{
      colspec = {lp{2.2cm}p{2.2cm}p{2.2cm}p{2.2cm}p{2.2cm}},
    }
    \toprule
           & \SetCell[c=2]{c} Training & & \SetCell[c=3]{c} Testing & & \\
    \cmidrule[lr]{2-3}\cmidrule[lr]{4-6}
            & KiTS & Radboudumc \newline training set & Radboudumc \newline test set & TCGA-KIRC  & Charité \newline Universitäts- \newline medizin Berlin \\
    \midrule
        Country & USA & Netherlands & Netherlands & USA & Germany \\
        Number of scans & 489 & 215 & 50 & 28 & 1510 \\
        Number of lesions & 2.7 $\pm$ 2.8 & 3.0 $\pm$ 5.0 & 3.9 $\pm$ 2.8 & 3.3 $\pm$2.6 &  2.0 $\pm$ 3.0 \\
        Lesion volume ($\times10^3$ mm$^3$) & 101 $\pm$ 236 & 17 $\pm$ 84 & 15 $\pm$ 47  & 136 $\pm$ 185 & 118 $\pm$ 218 \\
        Kidney volume ($\times10^3$ mm$^3$) & 278 $\pm$ 153 & 182 $\pm$ 104 & 132 $\pm$ 64 & 268 $\pm$ 158 & 277 $\pm$ 232 \\
        In-plane resolution (mm) & 0.80 $\pm$ 0.11 & 0.75 $\pm$ 0.07 & 0.76 $\pm$ 0.06 & 0.77 $\pm$ 0.08 & 1.0 $\pm$ 0.0 \\
        Slice thickness (mm) & 3.35 $\pm$ 1.70 & 1.22 $\pm$ 0.37 & 1.17 $\pm$ 0.38 & 2.81 $\pm$ 1.48 & 1.0 $\pm$ 0.0 \\ 
    \bottomrule
    \end{booktabs}
    \label{tab:training_data_characteristics}
    \end{table*}

    \subsection{Training Datasets}\label{training-datasets}
    \subsubsection{KiTS23 public training dataset}
    The 2023 Kidney Tumor Segmentation challenge training dataset consists of 489 CT scans coming from patients who underwent cryoablations, partial nephrectomy or radical
    nephrectomy for suspected renal cancer between 2010 and 2022 at a medical center in the United States of America \citep{heller_2023}. The most recent contrast-enhanced preoperative scan in either the corticomedullary
    or nephrogenic phase was selected and annotated by a pool of experts, trainees and lay-people using the following
    labels: kidney, tumor and cyst. The kidney region includes all
    parenchyma and the non-fat tissue within the hilum. The tumor
    regions comprises of masses found on the kidney that were suspected of
    being malignant and the cyst regions comprise of masses that were
    radiologically or pathologically determined to be cysts. The average age of the patients in the dataset is 59 ($\pm$ 14) years, with 37\% identifying as female. 91\% of the tumors is histologically confirmed to be malignant. 
    
    \subsubsection{Radboudumc public dataset}
    The KiTS23 dataset is derived solely from U.S.-based sources, limiting its generalizability. To enhance model robustness, we incorporated additional external data from the publicly available Kidney Abnormality dataset \citep{humpire-mamani_2023, mamani_2023}, collected at Radboudumc Nijmegen in the Netherlands. This dataset contains in total 215 contrast-enhanced thorax-abdomen CT scans from oncology patients. Out of the 215 scans, 113 cases are without lesions in the kidney, and 102 cases are with one or more kidney lesions, including cysts, lesions, masses, metastases, and tumors. Seventeen patients underwent left nephrectomy, and eighteen underwent right nephrectomy. The kidney region was annotated by radiologists and medical students, encompassing the renal cortex, medulla, and pyramid, while the renal mass region included only masses connected to the kidney parenchyma. Lesions in the collecting system were excluded. The patient cohort contains 44\% of patients identifying as female, with an average age of 60 years, ranging from 22 to 84 years \citep{mamani_2023}. 

    \subsubsection{Merging of training datasets}
    To combine the KiTS23 dataset with the Kidney Abnormality dataset from
    Radboudumc (hereafter referred to as the Radboudumc training dataset), a few adjustments were made. The primary difference between the datasets lies in their annotation protocols, for full annotation protocols see \citep{heller_2021, heller_2023, mamani_2023}. The KiTS23 dataset distinguishes between cysts and tumors, while in the Radboudumc dataset these categories were combined. To ensure consistency across datasets, the cyst and tumor masks in the KiTS23 dataset were merged into one category, aligning with the Radboudumc dataset. Besides practical challenges, the clinical boundary between renal tumors and cysts is inherently ambiguous. Kidney lesions include malignant solid tumors, benign solid masses (e.g., oncocytoma) \citep{kang_2014}, and cystic masses ranging from simple cysts (Bosniak categories I and II) to complex lesions with malignant potential (Bosniak categories III and IV) \citep{silverman_2019}. In this study, we therefore define the semantic segmentation task of background, kidney, and renal masses.  

\begin{figure}[h!]
        \centering
        \includegraphics[width=0.9\linewidth]{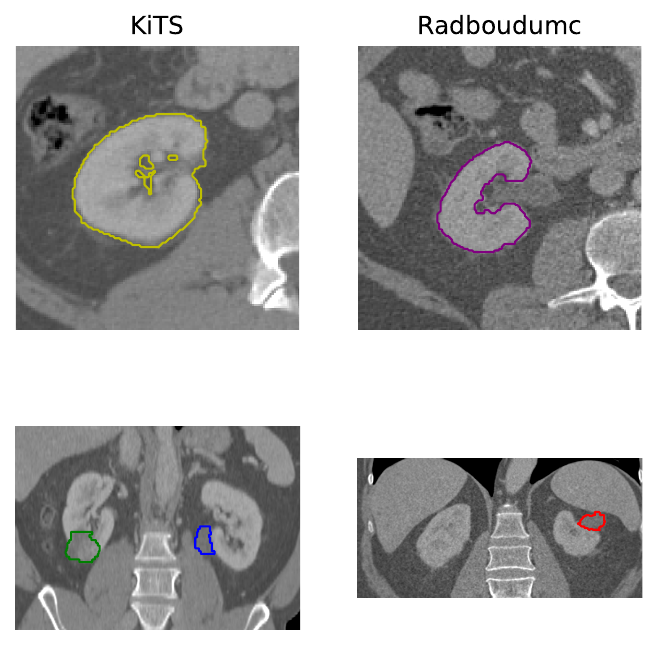}
        \caption{Visualization of the annotation protocols of the two
    training datasets. On the left, a segmentation mask from the KiTS
    dataset is shown in \textcolor{yellow}{\rule[-0.3mm]{.3cm}{.3cm}}, which shows that the hilum is included in the
    annotation. On the right, a segmentation mask from the Radboudumc
    dataset is shown in \textcolor{purple2}{\rule[-0.3mm]{.3cm}{.3cm}}, which shows that the hilum was left out of the kidney
    region while annotating. In the second row, we visualize that KiTS annotations distinguish between cysts \textcolor{blue2}{\rule[-0.3mm]{.3cm}{.3cm}} and tumors \textcolor{green2}{\rule[-0.3mm]{.3cm}{.3cm}} and Radboudumc annotations only include an abnormal mass region \textcolor{red}{\rule[-0.3mm]{.3cm}{.3cm}}.}
        \label{fig:annotation-protocol}
    \end{figure}

    Secondly, the annotation protocols for the kidney region differ between the two datasets: KiTS includes the hilum as part of the kidney region, whereas Radboudumc excludes it. An example illustrating these differences is provided in Figure~\ref{fig:annotation-protocol}. We decided not to re-annotate either of the kidney regions.
    
    \subsection{Test datasets}\label{external-validation}
    The robustness of the proposed deep learning framework was evaluated using one publicly available dataset and two proprietary datasets.  Table~\ref{tab:training_data_characteristics} presents the characteristics of the datasets and compares them to the training set. 

    \subsubsection{Radboudumc private test set}
    Alongside the publicly available Radboudumc kidney abnormality dataset, a separate private test set was collected. 
    This set consists of 20 cases without renal masses and 30 cases with renal masses. The set is annotated following the same protocol used for the public dataset. Additionally, to enable assessment of inter-reader variability, an independent observer (an experienced medical student) also segmented all cases in this private set. This inter-observer annotation facilitates evaluating the model’s performance relative to human observer variability.  The patient characteristics distribution is the same as reported for the Radboudumc training set.

    \subsubsection{TCGA-KIRC}
    Under the AIMI Annotations initiative, a subset of The Cancer Genome Atlas Kidney Renal Clear Cell Carcinoma Collection (TCGA-KIRC) \citep{akin_2016} is annotated for kidney,
    tumor and cyst regions, first by a trained AI model and corrected by an
    experienced radiologist \citep{murugesan_2024}. This set contains 28 valid imaging-mask pairs all from patients with clear cell renal cell carcinoma, incomplete imaging-mask pairs were excluded from analysis. The cohort is 41\% female, with a mean age of 57 ($\pm$ 10) years. 

    \subsubsection{Charité Universitätsmedizin Berlin private test set}
    The second proprietary dataset originates from Charité Universitätsmedizin Berlin, Germany. It comprises 1510 preoperative CT scans from 917 patients who underwent nephrectomy due to suspected renal tumors. The dataset includes scans from the early venous (n=764), delayed venous (n=220), and arterial (n=526) contrast phases.
    Each scan is accompanied by annotations for one kidney, including tumors and cysts, performed by a medical student and two radiologists with five and four years of experience, respectively. The contra-lateral kidney and metastases are not annotated.
    The dataset includes the following tumor subtypes: clear cell renal cell carcinoma (ccRCC, n=1069), papillary renal cell carcinoma (pRCC, n=216), chromophobe renal cell carcinoma (chrRCC, n=74), renal oncocytoma (RO, n=100), and other rare tumor subtypes (n=51). The patient cohort is 30\% female, with a mean age of 63 ($\pm$ 11) years.
    
    \subsection{Deep learning framework}\label{deep-learning-framework}
    The deep learning method used in this paper is the nnU-Netv2
    framework \citep{isensee_2021, isensee_2024}. The nnU-Net framework eliminates the need for model architecture tweaking and hyperparameter tuning. The framework expects a supervised training dataset, and based on fixed parameters, rule-based parameters and empirical parameters the optimal network architecture and training parameters are determined. In this research, we used the newly introduced Residual Encoder (ResEnc) presets \citep{isensee_2024}, in particular the L(arge) version. Both full-resolution and low-resolution configurations were trained, along with a cascade model. Cross-validation results indicated that the full-resolution model performed best for this task. The resulting final model setup is presented in Table \ref{tab:nnunet-setup}.

    The model (named Renal-Net) is made available in two ways, by making the model weights and the code to run the model accessible on Zenodo(\url{https://doi.org/10.5281/zenodo.15315330}) and Github (\url{https://github.com/DIAGNijmegen/oncology-kidney-abnormality-segmentation}) and by providing a ready to use algorithm on \url{grand-challenge.org} (\url{https://grand-challenge.org/algorithms/kidney-abnormality-segmentation/}). 

    \subsubsection{Postprocessing}\label{postprocessing}
    After the prediction is made by the nnU-Net algorithm, a postprocessing
    pipeline is executed. 
    Predicted renal mass regions that are not attached to the kidney region were removed. However, if the volume of the predicted lesion is bigger than 100 cm$^3$, it was kept. We noticed that large tumors can suppress the kidneys to such an extend that the model does not recognize it as kidney anymore. This threshold is based on the average kidney volume, measured on ultrasound imaging, 146 cm$^3$ for the left kidney and 134 cm$^3$ for the right kidney \citep{emamian_1993}. For predicted masses attached to the kidney, only those with an axial diameter greater than three mm were retained. According to the RECIST criteria \citep{eisenhauer_2009}, lesions smaller than ten mm are not considered target lesions, and those smaller than five mm are regarded as non-measurable. A threshold of three mm was chosen to allow the inclusion of some very small lesions while maintaining a low rate of false positives.

    \section{Experiments}\label{experiments}
    Model training was done on a single NVIDIA A100 40 GB GPU. In our experiments, we used the ensemble model of the 5-fold cross validation training. 
    
    \subsection{Evaluation}
    The proposed model (Renal-Net) is evaluated on multiple grounds. First, we evaluated the segmentation and detection performance on the three independent test datasets comparing the model with and without post-processing.
    Secondly, for the TCGA-KIRC and Charité Universitätsmedizin Berlin test sets we performed subgroup analysis to assess potential performance variations.
    
    Performance is reported for three regions: (1) kidney only, (2) kidney + masses, and (3) renal masses only. This structured evaluation enables a clear comparison, isolating kidney segmentation performance - particularly relevant for comparison with TotalSegmentator, which was trained on healthy kidneys - and assessing our model's robustness in segmenting kidneys without lesions. Following the KiTS challenge evaluation \citep{heller_2023}, performance metrics are also reported on the combined segmentation of kidneys and lesions.

    \subsubsection{Reference models}
    To benchmark Renal-Net, we compared it to two publicly available models, TotalSegmentator \citep{wasserthal_2023} and the BAMF model \citep{murugesan_2024}, which is the only publicly available model specifically trained for renal mass segmentation. The BAMF model, similar to the proposed model, is trained primarily on CT.
    
    TotalSegmentator \citep{wasserthal_2023} provides separate labels for right and left kidney as well as for cysts in each kidney. To align with our evaluation approach, we combined right and left kidney regions into one kidney region and right and left kidney cysts into renal mass region.
    
    Similar to the proposed model, the BAMF model is nnU-Net-based and specifically trained for kidney tumor segmentation. The training dataset includes the KiTS dataset and the TCGA-KIRC dataset. For the latter, annotations were obtained via active learning, where model predictions on unlabeled cases were expert-corrected and iteratively added for retraining.

    \subsubsection{Radboudumc test set}
    We compared the performance of our proposed model with TotalSegmentator \citep{wasserthal_2023} and the BAMF model \citep{murugesan_2024}  on the Radboudumc test sets.
    The Radboudumc private test dataset contains two subsets, B20 with 20 patients who have healthy kidneys, and B30 with 30 patients who have at least one lesion in the kidney. Performance is reported on the two subsets separately. 
    
    \subsubsection{TCGA-KIRC test set}
    To evaluate the proposed model on the TCGA-KIRC test set, we compared its performance against TotalSegmentator.
    We excluded the BAMF model from this comparison because TCGA-KIRC is part of its training set, which would bias the evaluation.
    
    \subsubsection{Charité Universitätsmedizin Berlin test set}
    Due to the relatively large size of the Charité Universitätsmedizin Berlin test set and the previously observed poor performance on the renal mass region on the other two test sets, TotalSegmentator was excluded from this analysis. 
    
    In our proposed model, segmentation was performed for both the left and right kidney. However, for the Charité dataset, annotations were available for only a single kidney per scan, specifically the kidney with a lesion. To address this, we limited our analysis to the annotated kidney. Specifically, overlap between each predicted region and the available annotation was assessed (overlap was considered if at least one voxel overlapped) and only the region that overlapped with the reference was selected for evaluation. If no overlap was found, an evaluation score of zero was assigned.

    \subsection{Statistical tests}
    To investigate whether the proposed model significantly outperforms baseline methods in terms of Dice score, the Mann-Whitney U test was performed. The Mann-Whitney U test was selected over rank-based alternatives, as it operates on the full score distributions rather than average ranks, providing greater statistical power for pairwise superiority testing. We tested the one-sided alternative hypothesis that the proposed model achieves superior Dice scores relative to each comparison model. We only performed the significance test for the Dice metric, as this is our main performance metric. A hierarchical statistical analysis plan was used, where we first tested for superiority over TotalSegmentator and if this condition is met we tested for superiority over the BAMF-model. Since TotalSegmentator is not trained specifically for kidney lesions, significance testing was not performed for this comparison. We corrected for multiple testing by using the Holm-Bonferroni adjustment for the multiple datasets and segmentation regions we tested for each model comparison. A significance threshold ($\alpha$) of 0.05 was used.

    \subsection{Segmentation performance}
    The Dice metric was used as overlap-based metric to compare with previous research but a boundary-based metric was added as is recommended in the Metrics Reloaded paper \citep{maier-hein_2024}. For boundary-based metrics we report Hausdorff distance at 95 percentile.

    \subsection{Detection performance}
    Detection performance of renal masses was assessed by reporting false-positive and false-negative findings of the model. A renal mass was considered successfully detected if the overlap between the predicted region and the annotated region exceeded a specified threshold. In this study, an overlap threshold of 0.5 was used, with additional results at a threshold of 0 provided in Appendix~\ref{app:detection}. The overlap is defined as the intersection over union. In this study, the main focus was segmentation performance and detection performance is derived from segmentation predictions. We therefore focus on precision, recall and F1-scores rather than doing FROC analysis.
    
    \subsection{Subgroup analysis}
    We performed subgroup analyses for the TCGA-KIRC and Charité Universitätsmedizin Berlin test datasets to identify potential biases in model performance. The Radboudumc test set was excluded from this analysis due to the absence of detailed patient-level characteristics We investigated the robustness across different subgroups of patient sex, patient age, contrast phase of the scan and tumor histologic subtype. 

\section{Results} \label{results}
Segmentation and detection outcomes are reported separately, with segmentation performance provided for each anatomical region. Significance test are performed solely for the Dice metric. In Appendix~\ref{app:segmentation}, we present all segmentation metrics for each test set in tables.

\subsection{Segmentation results}

\subsubsection{Healthy kidneys}
\begin{figure}[h]
    \centering
    \includegraphics[width=\linewidth]{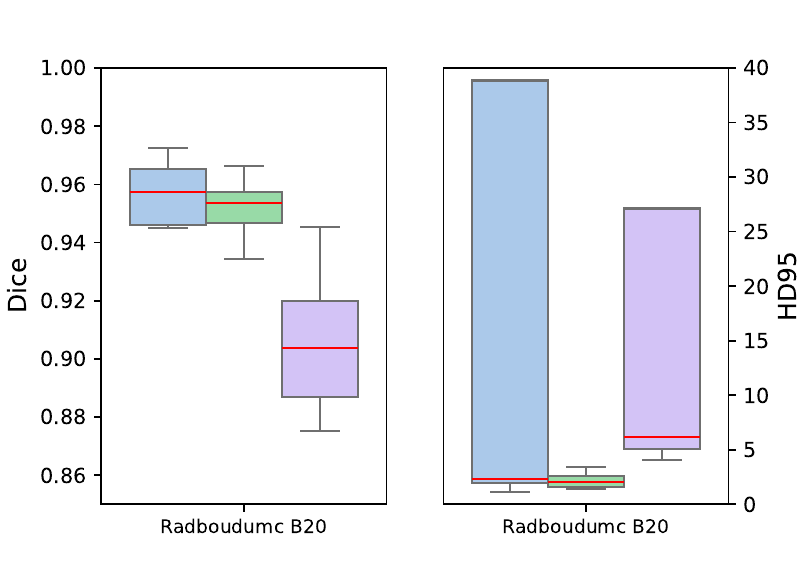}
    \caption{Segmentation results on the healthy cohort of the Radboudumc test set expressed in Dice (left) and Hausdorff distance (mm) at 95th percentile (right). We present Renal-Net (ours) \textcolor{blue1}{\rule[-0.3mm]{.3cm}{.3cm}}, TotalSegmentator \textcolor{green1}{\rule[-0.3mm]{.3cm}{.3cm}} and the BAMF model \textcolor{purple1}{\rule[-0.3mm]{.3cm}{.3cm}}. The boxplots show the median (horizontal red line in the box), quartiles (presented by the box) and the full distribution (presented by the whiskers). For the purpose of readability we omitted the outliers from these plots.}
    \label{fig:B20}
\end{figure}

The Radboudumc private test set contains a subset (B20) with patients with healthy kidneys. Figure~\ref{fig:B20} presents the results of the proposed model (Renal-Net) and the reference models on this test dataset. The inter-observer variability, as reported by an independent human observer with a mean Dice of 0.94 ($\pm$0.01) on the B20 subset \citep{mamani_2023}, is comparable to the performance achieved by the proposed model (0.93 $\pm$ 0.06) and TotalSegmentator (0.95 $\pm$ 0.01), while the BAMF model achieves 0.90 $\pm$ 0.02. The proposed model does not significantly outperform TotalSegmentator (corrected p-value $>$ 0.05) based on Dice on the healthy kidney subset, therefore superiority over the BAMF-model is not tested. Based on the performance measured in HD95, TotalSegmentator and the proposed model achieve mean scores of 7.39 $\pm$ 23.71 and 55.80 $\pm$ 101.21 respectively. From the boxplots in Figure~\ref{fig:B20} it can be observed that the median HD95 of the proposed model is close to the median HD95 of TotalSegmentator, however the quartiles differ substantially indicating a few cases where false positives for structures physically far from the kidney region are segmented.

\subsubsection{Kidney region}
\begin{figure}[h!]
    \centering
    \includegraphics[width=\linewidth]{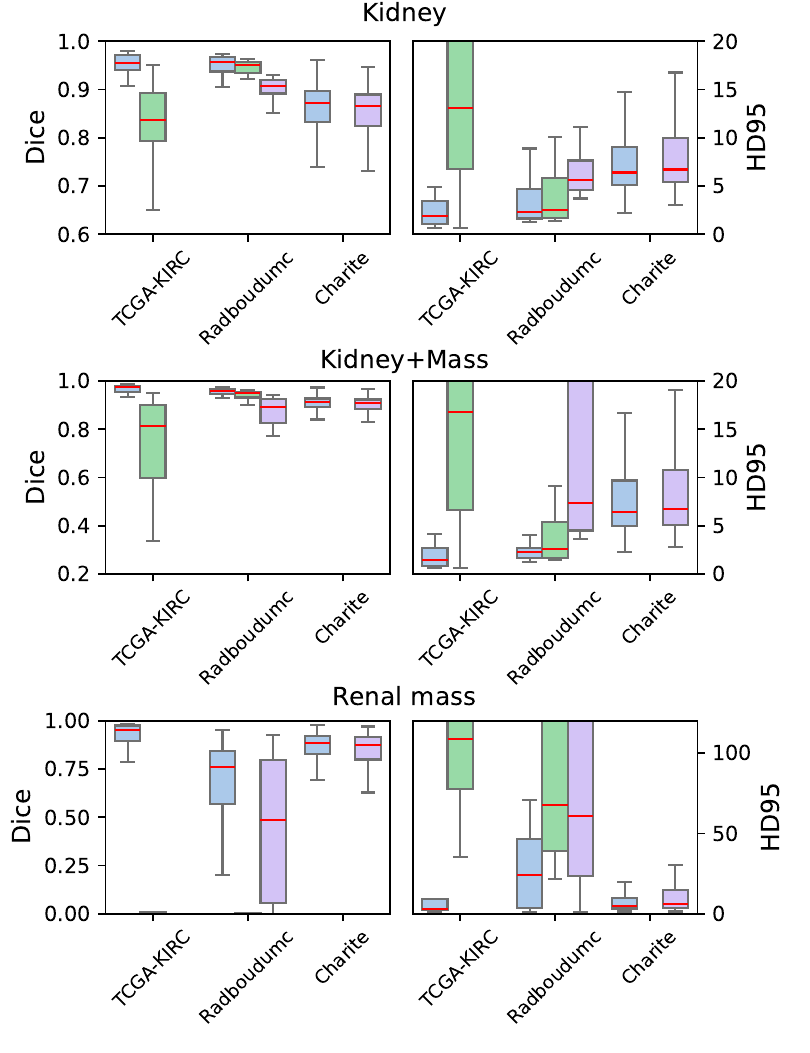}
    \caption{Segmentation results expressed in Dice (left column) and Hausdorff distance (mm) at 95th percentile (right column). The results are shown for three segmentation regions, Kidney, Kidney+Mass and Renal mass. We present results per dataset (excluding the patients with healthy kidneys in the Radboudumc test set) and show Renal-Net (ours) \textcolor{blue1}{\rule[-0.3mm]{.3cm}{.3cm}}, TotalSegmentator \textcolor{green1}{\rule[-0.3mm]{.3cm}{.3cm}} and the BAMF-model \textcolor{purple1}{\rule[-0.3mm]{.3cm}{.3cm}}. The boxplots show the median (horizontal red line in the box), quartiles (presented by the box) and the full distribution (presented by the whiskers). For the purpose of readability we omitted the outliers from these plots. The BAMF-model is excluded in the TCGA-KIRC analysis due to training data overlap and TotalSegmentator is excluded from the Charité analysis due to observed poor performance on the other two test sets.}
    \label{fig:segmentation-results}
\end{figure}

The first row in Figure~\ref{fig:segmentation-results} presents the segmentation results for the kidney region. Median Dice scores for all models across the three datasets are around 0.90. The proposed model achieved the highest mean Dice scores on both the TCGA-KIRC (0.95 $\pm$ 0.03) and Radboudumc (0.94 $\pm$ 0.05) datasets, outperforming TotalSegmentator (0.80 $\pm$ 0.14 and 0.92 $\pm$ 0.10). On the Radboudumc dataset, the proposed model also outperformed the BAMF model (0.90 $\pm$ 0.04). Inter-observer variability yielded mean Dice scores of 0.925 ($\pm$0.051) on this test set \citep{mamani_2023}. 

For HD95, the proposed model demonstrated substantially lower values on the TCGA-KIRC dataset (3.79 $\pm$ 5.89) compared to TotalSegmentator (25.90 $\pm$ 32.41). A similar trend as for the healthy kidneys was observed on the Radboudumc test set, where the proposed model segments for a few cases structures physically far from the kidneys resulting in a higher HD95 score (44.32 $\pm$ 99.98).

On the Charité Universitätsmedizin Berlin private test set, the proposed model and BAMF model reached comparable performance, both in terms of mean Dice (0.85 $\pm$ 0.10 and 0.83 $\pm$ 0.13, respectively) and HD95 (9.84 $\pm$ 16.11 and 12.19 $\pm$ 18.90, respectively).

Across all datasets for the kidney region, the proposed model yielded significantly higher Dice scores than the reference models (corrected p-values $<$ 0.001).

\subsubsection{Kidney+Mass region}
The second row in Figure~\ref{fig:segmentation-results} shows the results on the kidney+mass region. We see very similar trends as for the kidney region alone. The proposed model outperforms TotalSegmentator on the TCGA-KIRC dataset. On the Radboudumc test set, the BAMF model underperforms compared to both the proposed model and TotalSegmentator. Inter-observer variability yielded mean Dice scores of 0.941 ($\pm$0.017) on the Radboudumc test set \citep{mamani_2023}, where the proposed model achieved 0.95 ($\pm$0.02). The proposed model and the BAMF model performed similarly in terms of Dice on the Charité Universitätsmedizin Berlin private test set. Across all datasets for the kidney+mass region, the proposed model achieved significantly higher Dice scores than the reference models with corrected p-values $<$ 0.001.

The mean HD95 scores as presented in Appendix~\ref{app:segmentation} highlight that the proposed model (43.66 $\pm$ 100.00) and the BAMF model (70.58 $\pm$ 88.95) segment for a few cases regions physically far from the kidney region, while TotalSegmentator (15.40 $\pm$ 37.83) does not show the same pattern.

\subsubsection{Renal mass region}
Lastly, the renal mass results are presented in the bottom row of Figure~\ref{fig:segmentation-results}. Both the Dice and HD95 scores are more widely spread, but the proposed model consistently outperforms TotalSegmentator and the BAMF model. TotalSegmentator performs much worse on the TCGA-KIRC dataset compared to the proposed model both on the mean Dice (0.05 $\pm$ 0.13 and 0.86 $\pm$ 0.25, respectively) and mean HD95 metric (120.18 $\pm$ 61.30 and 14.94 $\pm$ 25.81, respectively). On the Radboudumc test set, the proposed model and the BAMF-model both have varying performance on the renal mass region when looking at the mean Dice score (0.66 $\pm$ 0.29 and 0.45 $\pm$ 0.35 respectively), although the proposed model on average scores better than the BAMF-model. Next to that, the HD95 metric shows superior performance of the proposed model. Inter-observer variability yielded a mean Dice score of 0.664 ($\pm$0.274) for the renal mass region \citep{mamani_2023}.

Similarly to the other tested segmentation regions, the performance of the proposed model and the BAMF model on the Charité Universitätsmedizin Berlin private test set is close together. Measured in mean Dice the proposed model and the BAMF model reach 0.83 $\pm$ 0.17 and 0.81 $\pm$ 0.19 respectively, in HD95 the models reach 10.75 $\pm$ 16.86 and 13.43 $\pm$ 18.47 respectively. To investigate the differences in performance between the models one step further we compare the 5th percentile, which focuses on the difficult cases. These results are presented in Table~\ref{tab:quantile}. We can see that the BAMF-model is under performing compared to the proposed model on all regions but especially the renal mass region (5th percentile Dice of 0.53 and 0.38 for the proposed and BAMF-model). By looking at the 95th percentile we can see that this advantage is not followed by worse performance on the easy cases (both models achieve 0.95 Dice on the 95th percentile).

The proposed model achieved significantly higher Dice scores than the reference models across all datasets for the renal mass region with corrected p-values $<$ 0.001.

\begin{table}[!h]
\small
\centering
\caption{Quantile investigation of the model performance on the Charité Universitätsmedizin Berlin test set. Dice scores are reported at the 5th and 95th percentiles to characterize the distribution of segmentation performance across patients, capturing both typical failure cases and upper performance bounds. PP = post-processing, best values per metric per region are presented in bold.}
\begin{booktabs}{
  colspec = {lp{1.8cm}p{1.8cm}},
}
\toprule
       Kidney & Dice 5th \newline percentile & Dice 95th \newline percentile \\
\midrule
    BAMF-model & 0.65 & 0.91 \\
    Renal-Net (ours) & 0.71 & 0.92 \\
    Renal-Net + PP (ours) & \textbf{0.71} & \textbf{0.92} \\
\midrule
       Kidney+mass &  & \\
\midrule
    BAMF-model & 0.78 & 0.94 \\
    Renal-Net (ours) & 0.84 & 0.95 \\
    Renal-Net + PP (ours) & \textbf{0.84} & \textbf{0.95} \\
\midrule 
       Renal Mass &  &  \\
\midrule
    BAMF-model & 0.39 & 0.95 \\
    Renal-Net (ours) & 0.53 & 0.95 \\
    Renal-Net + PP (ours) & \textbf{0.53} & \textbf{0.95} \\ 
\bottomrule
\end{booktabs}
\label{tab:quantile}
\end{table}

\subsection{Detection results}
Table~\ref{tab:detection} presents the detection results of the proposed model and the reference models on all test sets. We present the detection results using an overlap threshold of 0.5, based on intersection over union, the results for a lower threshold can be found in Appendix~\ref{app:detection}. The proposed model outperforms the reference models on all three metrics on all three test datasets. The difference in scores between the datasets highlights the different nature of the three test datasets.

\begin{table*}[t]
\centering
\small
\caption{Detection performance measured in precision, recall and F1-score, where an overlap threshold of 0.5 is used to determine if a lesion is detected. We present the averages and standard deviations for all models and all three test datasets. PP = post-processing, best values per metric per dataset are presented in bold.}
\begin{booktabs}{
  colspec = {p{4cm}p{2.5cm}p{2.5cm}p{2.5cm}},
}
\toprule
       Radboudumc & Precision & Recall & F1-score  \\
\midrule
    TotalSegmentator & 0.00 ($\pm$ 0.00) & 0.00 ($\pm$ 0.00) & 0.00 ($\pm$ 0.00) \\
    BAMF-model & 0.34 ($\pm$ 0.28) & 0.44 ($\pm$ 0.35) & 0.34 ($\pm$ 0.24) \\
    Renal-Net (ours) & 0.56 ($\pm$ 0.35) & 0.52 ($\pm$ 0.34) & 0.51 ($\pm$ 0.30) \\
    Renal-Net + PP (ours) & \textbf{0.65 ($\pm$ 0.30)} & \textbf{0.52 ($\pm$ 0.34)} & \textbf{0.59 ($\pm$ 0.27)} \\
\midrule
       TCGA-KIRC & Precision & Recall & F1-score \\
\midrule
    TotalSegmentator & 0.00 ($\pm$ 0.00) & 0.00 ($\pm$ 0.00) & 0.00 ($\pm$ 0.00) \\
    Renal-Net (ours) & 0.75 ($\pm$ 0.28) & 0.56 ($\pm$ 0.33) & 0.64 ($\pm$ 0.27) \\
    Renal-Net + PP (ours) & \textbf{0.75 ($\pm$ 0.28)} & \textbf{0.56 ($\pm$ 0.33)} & \textbf{0.64 ($\pm$ 0.27)} \\
\midrule 
       Charité Universitäts- \newline medizin Berlin & Precision & Recall & F1-score   \\
\midrule
    BAMF-model & 0.68 ($\pm$ 0.34) & 0.79 ($\pm$ 0.34) & 0.70 ($\pm$ 0.32) \\
    Renal-Net (ours) & 0.76 ($\pm$ 0.32) & 0.83 ($\pm$ 0.30) & 0.77 ($\pm$ 0.29) \\
    Renal-Net + PP (ours) & \textbf{0.77 ($\pm$ 0.31)} & \textbf{0.83 ($\pm$ 0.31)} & \textbf{0.78 ($\pm$ 0.29)} \\
\bottomrule
\end{booktabs}
\label{tab:detection}
\end{table*}

\subsection{Subgroup analysis}
\begin{figure}[h]
    \centering
    \includegraphics[width=\linewidth]{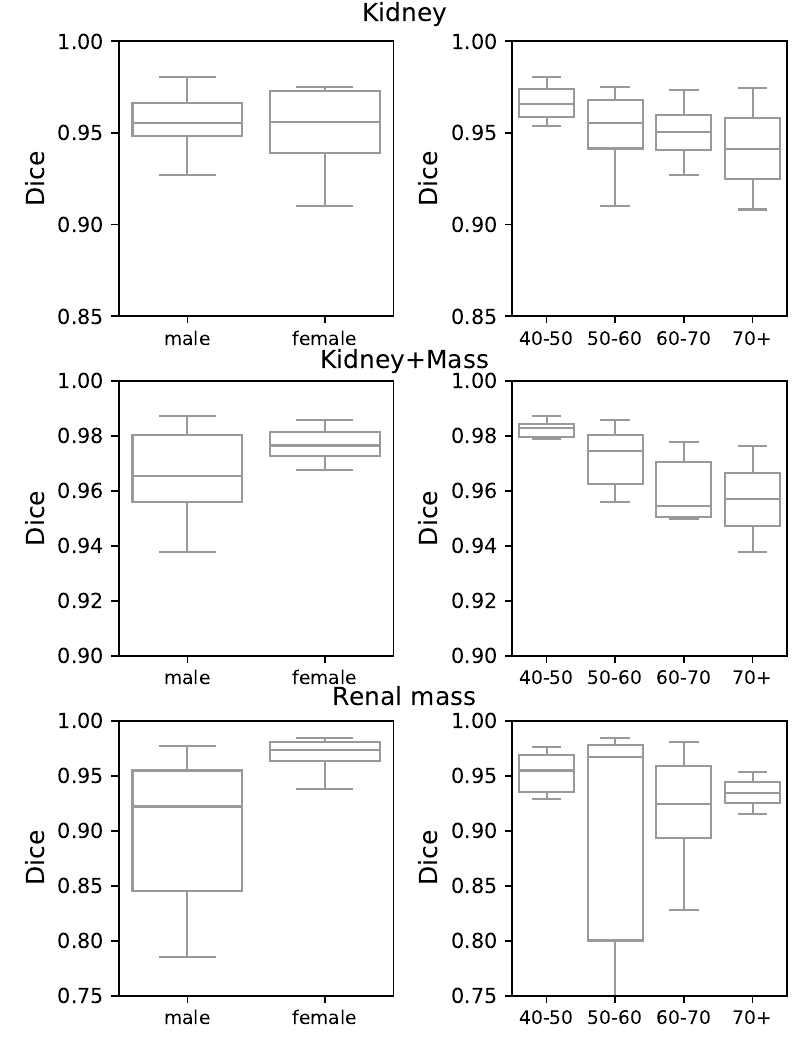}
    \caption{Subgroup analysis performed on the TCGA-KIRC test set. Subgroups that are tested for are patient sex and patient age. We show performance measured in Dice. The boxplots show the median (horizontal line in the box), quartiles (presented by the box) and the full distribution (presented by the whiskers). For the purpose of readability we omitted the outliers from these plots.}
    \label{fig:subgroup-tcga-kirc}
\end{figure}

\begin{figure}[h]
    \centering
    \includegraphics[width=\linewidth]{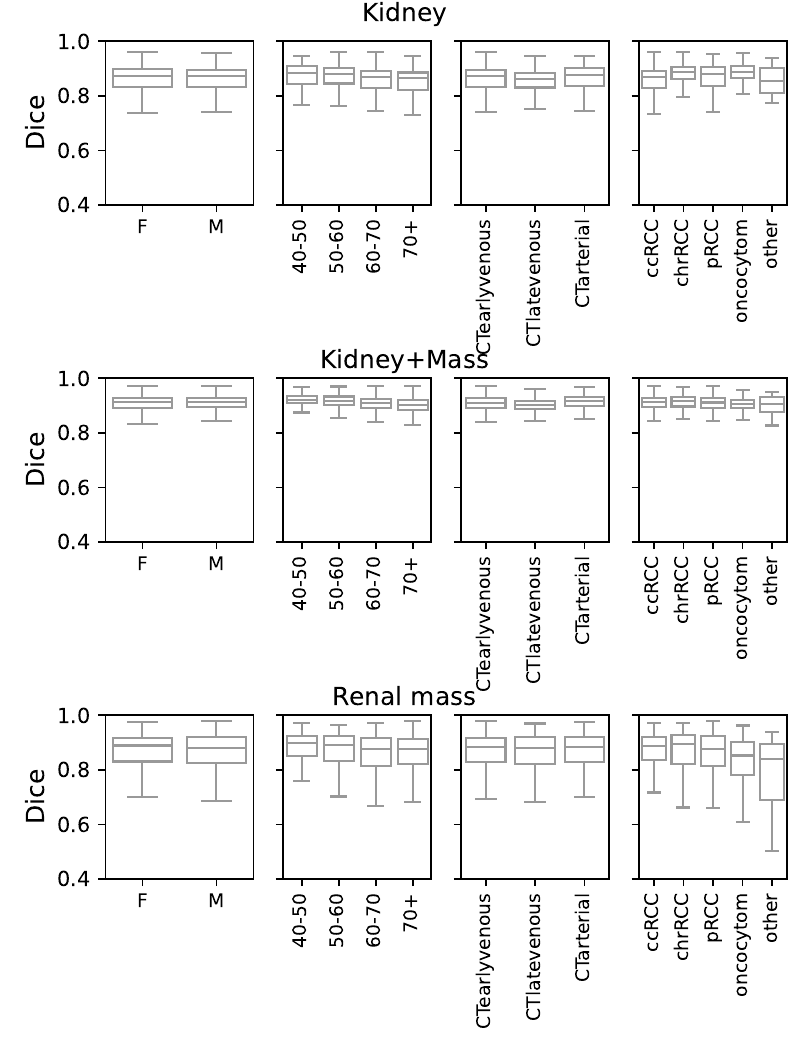}
    \caption{Subgroup analysis performed on the Charité Universitätsmedizin Berlin test set. Subgroups that are tested for are patient sex, patient age, CT contrast phase and tumor histologic subtype. We show performance measured in Dice. The boxplots show the median (horizontal line in the box), quartiles (presented by the box) and the full distribution (presented by the whiskers). For the purpose of readability we omitted the outliers from these plots.}
    \label{fig:subgroup-charite}
\end{figure}

Subgroup analysis was performed on the TCGA-KIRC and Charité Universitätsmedizin Berlin private test sets using the proposed model (results of the BAMF-model can be found in Appendix~\ref{app:subgroup-analysis}). We investigated the robustness across different subgroups of patient sex, patient age, contrast phase of the scan and tumor histologic subtype. Figure~\ref{fig:subgroup-tcga-kirc} presents the results for the TCGA-KIRC test set, Figure~\ref{fig:subgroup-charite} presents the results for the Charité test set. An additional subgroup analysis of the BAMF model on the Charité test set can be found in Appendix~\ref{app:subgroup-analysis}.

\subsection{Qualitative results}
\begin{figure*}[h]
    \centering
    \includegraphics[width=0.8\linewidth]{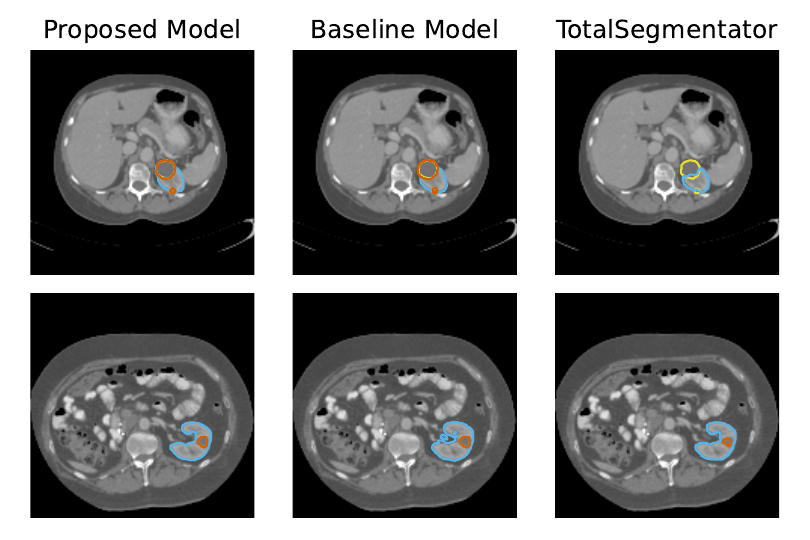}
    \caption{Presented are two cases (top and bottom) from the Radboudumc test set and the predictions of the proposed model, the BAMF-model (baseline) and TotalSegmentator. The reference renal mass mask \textcolor{quali_yellow}{\rule[-0.3mm]{.3cm}{.3cm}} and the renal mass predictions \textcolor{quali_orange}{\rule[-0.3mm]{.3cm}{.3cm}} are shown. The predicted kidney \textcolor{quali_blue}{\rule[-0.3mm]{.3cm}{.3cm}} is also highlighted. The biggest lesion in the top row presents as a homogeneously fluid-filled, non-enhancing mass consistent with a simple or low-complexity cyst.}
    \label{fig:qualitative-gabriel1}
\end{figure*}

In this section, we present qualitative results of the segmentations of the proposed and reference models. Figure~\ref{fig:qualitative-gabriel1} shows two cases from the Radboudumc test set where both the proposed and BAMF-model (baseline) are able to detect the renal mass (reference mask in yellow, predictions in orange), while TotalSegmentator misses the lesions in the first case. The biggest lesion in the top row of Figure~\ref{fig:qualitative-gabriel1} presents as a homogeneously fluid-filled, non-enhancing mass consistent with a simple or low-complexity cyst. The segmentation of the renal mass as predicted by the proposed model more closely mimics the annotation mask compared to the segmentation predicted by the BAMF-model (baseline). The kidneys (blue) are similar between all models. 

\begin{figure*}[h]
    \centering
    \includegraphics[width=0.7\linewidth]{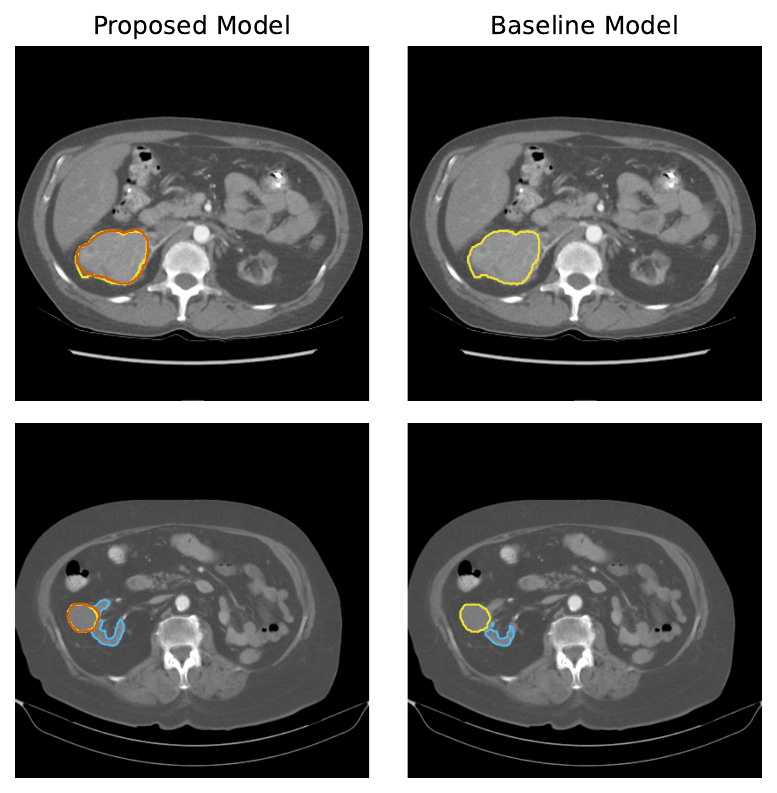}
    \caption{Presented are two cases (top and bottom) from the Charité Universitätsmedizin Berlin test set and the predictions of the proposed model and the BAMF-model (baseline). The reference renal mass mask \textcolor{quali_yellow}{\rule[-0.3mm]{.3cm}{.3cm}} and the renal mass predictions \textcolor{quali_orange}{\rule[-0.3mm]{.3cm}{.3cm}} are shown. The predicted kidney \textcolor{quali_blue}{\rule[-0.3mm]{.3cm}{.3cm}} is also highlighted.}
    \label{fig:charite_case2}
\end{figure*}

Figure~\ref{fig:charite_case2} presents two cases from the Charité Universitätsmedizin Berlin test set. The papillary renal cell carcinoma in the first row (yellow ground truth) has grown around the kidney and the characteristic kidney shape is not detectable anymore. The second row shows a tumor case on a remnant kidney. The proposed model did segment the tumors (orange) in both cases and did correctly capture the remnant kidney (blue). The BAMF-model did segment parts of the remnant kidney but could not correctly delineate the uncommon tumor shapes.

In Appendix~\ref{app:qualitative}, we present additional interesting cases on the Charité Universitätsmedizin Berlin test set. In Figure~\ref{fig:case1}, both models were able to correctly delineate the kidneys (blue) but failed to segment the clear cell renal cell carcinomas (yellow ground truth). In Figure~\ref{fig:case3}, both models correctly segment the kidneys (blue) and are able to localize the tumors (orange). However, the BAMF-model (baseline) was better in capturing the whole tumor. The scan in the upper row has visible horizontal lines through the scan, possibly caused by metal. In Figure~\ref{fig:case4}, we present a case where both models were able to correctly segment a transplanted kidney (blue) despite its uncommon orientation and location in the pelvic area. 

\section{Discussion} \label{Discussion}
In this research, we validated and investigated the robustness of a segmentation model (named Renal-Net) based on the state-of-the-art nnU-Net framework and trained on two publicly available training datasets. We tested the segmentation and detection performance on three test datasets. Two test datasets are external for the proposed model. Except for the healthy kidney subset, our model significantly outperforms the reference models. Additionally, on the Radboudumc private test set, the model's performance was within the range of human inter-observer variability, indicating that it performs comparable to an independent human annotator. Furthermore, these results suggest the model's predictions are realistic and clinically relevant. Our subgroup analysis showed that differences in performance between subgroups in sex, age, and tumor histologic subtypes are small and that the model is robust to use of different contrast phases. 

For the healthy kidney subset, TotalSegmentator achieves a lower median HD95 than the proposed model. We noticed that the proposed model predicted in a few cases structures physically far from the kidneys, which resulted in higher HD95 scores. In deployment, these structures could be flagged after connected component analysis or a region of interest crop could be made before or after the segmentation model using atlas based registration techniques \citep{buddenkotte_2024a}. While such shifts have less impact on volumetric Dice scores, they can noticeably affect HD95.

The detection performance of our proposed model is superior to the reference models, although it still generates some false positives and false negatives, as is reflected in the precision, recall and F1-score. Future work should focus on reducing these detection errors to better match the clinical requirements. Clinically, this segmentation
model could be used in multiple scenarios. One possible scenario is using the model as a standalone tool to assist radiologists directly. In this setting, the radiologist reviews the AI-generated segmentation and subsequently decides whether to accept or reject its predictions. False-positive detections could then be easily identified and dismissed by the radiologist. However, false-negative detections present a higher risk in this scenario, as they might incorrectly reassure the radiologist that no lesions are present in the kidneys, potentially leading to missed diagnoses. Another scenario of using this algorithm in clinical practice is to integrate it in a broader automated kidney diagnostic pipeline. The radiologist might not see the in-between results of the pipeline but only the outcome. In this case, both false negatives and false positives can negatively impact the outcomes of such workflows.

The qualitative results showed that the kidney segmentation performance is good for both the proposed and reference models. However, renal mass segmentation was for some cases better captured by the proposed model. The proposed model also showed more robust performance across a wide range of kidney anatomies.

In this study, we deliberately chose not to re-annotate either of the training datasets to standardize the inclusion or exclusion of the kidney hilum. We show preliminary evidence to support the hypothesis that exposure to annotation variability during training improves model robustness in real-world settings. By combining these heterogeneous datasets, our model is exposed to a broad range of anatomical variations, scanning protocols, and slice thicknesses, promoting robustness against distribution shifts. Clinically, achieving a high level of robustness is ultimately more valuable than minor incremental gains in segmentation accuracy. A potential application of our model in clinical practice includes evaluating kidney volumes and tracking relative volume changes over multiple time points. This can serve as a diagnostic marker to guide procedures such as active surveillance. Another interesting use case could be predicting the Mayo Imaging Classification score \citep{bais_2024} for patients with ADPKD.

Furthermore, we trained the model to detect and segment renal masses as a single merged class, rather than separating cystic and solid lesions. This decision was motivated by the considerable overlap in imaging characteristics between these subtypes, e.g., Bosniak IV cystic lesions may present with both septations and solid components \citep{silverman_2019}, making a strict class boundary ambiguous. Our results demonstrate that this approach is effective for renal mass detection and segmentation. Whether two separate models, one for cystic and one for solid lesions, could further improve performance remains an open question and represents a direction for future work.

Besides CT imaging, MR imaging is a widely used modality for renal imaging. The model presented in this work is trained and validated on CT imaging only. Recent work has shown that with some extra preprocessing steps, a CT-only model can be applied to MR imaging \citep{hantze_2025}. Although the results are not yet satisfactory, this setup could be used to accelerate annotating MRI for training an MRI specific model.

The proposed model has an important limitation, the model is not able to recognize kidneys and renal masses on CT scans without contrast. This problem could potentially be mitigated by retraining the model on a training dataset that does include non-contrast scans. While contrast phase information was available for one test dataset, contrast agent specifications were not recorded across sites. We acknowledge that a systematic analysis of contrast agent variability would have provided further insight into model generalizability. Nevertheless, the consistently high performance observed across all test datasets indirectly demonstrates that the model is robust to variability in scanner type, acquisition protocol, and other site-specific factors.
A further limitation is the incomplete metadata availability across test sets, which limited a fully consistent subgroup analysis. Analyses were conducted for two of the three test sets where demographic metadata was available. Future work with complete metadata across all test sets would enable a more comprehensive assessment of model fairness across demographic subgroups. Finally, we acknowledge that our validation does not include pediatric cases nor data from Africa, South-America, Asia or Australia. Future validation studies should be performed to further test the generalizability of the model.

\section{Conclusion} \label{Conclusion}
In this study, we validated our proposed renal mass segmentation model (Renal-Net) across three independent test datasets, demonstrating superior segmentation and detection performance compared to existing publicly available models. Through subgroup analyses, we confirmed that the model performs robustly across various patient subgroups and imaging conditions. Our results indicate that combining publicly available datasets with the state-of-the-art nnU-Net framework is sufficient for developing a robust and generalizable segmentation model for renal masses. We encourage researchers and clinicians to utilize our publicly available model and further evaluate its performance using imaging data from their own medical centers.


\acks{This research is funded by the European Union under HORIZON-HLTH-2022: COMFORT (101079894). Views and opinions expressed are however those of the author(s) only and do not necessarily reflect those of the European Union or European Health and Digital Executive Agency (HADEA). Neither the European Union nor the granting authority can be held responsible for them.}

%
\ethics{The work follows appropriate ethical standards in conducting research and writing the manuscript, following all applicable laws and regulations regarding treatment of animals or human subjects.}

\coi{Bram van Ginneken is a shareholder of Thirona BV and Plain Medical. Kiran Vaidhya Venkadesh is a shareholder of Plain Medical. The other authors declare that they do not have any conflicts of interest.}

\data{The data used for training the model are publicly available via \url{https://kits-challenge.org/kits23/} and \url{https://zenodo.org/records/8014290}. The TCGA test set is publicly available via \url{https://www.cancerimagingarchive.net/collection/tcga-kirc/} and \url{https://zenodo.org/records/13244892}. The test datasets from Radboudumc and Charité Universitätsmedizin Berlin were used under approval for the current study. Restrictions apply to the availability of these datasets and so they are not publicly available.}

\bibliography{bibliography}


\clearpage
\appendix
\section{nnU-Net configuration}
Table \ref{tab:nnunet-setup} presents the final training and architecture setup of the proposed model.

\begin{table}[ht]
\centering
\small
\caption{Overview of nnU-Net 3D full-resolution configuration used in this research. This configuration is determined by the nnU-Net framework and we did not adjust any settings.}
\begin{tabular}{p{4cm} p{4cm}}
\hline
\textbf{Category} & \textbf{Configuration} \\
\hline
\multicolumn{2}{l}{\textbf{Preprocessing}} \\
Data identifier & nnUNetPlans\_3d\_fullres \\
Preprocessor & DefaultPreprocessor \\
Patch size & [160, 224, 192] \\
Median image size & [390, 512, 512] \\
Spacing (mm) & [0.7695, 0.7695, 0.7695] \\
Normalization & CTNormalization \\
\hline
\multicolumn{2}{l}{\textbf{Network Architecture}} \\
Class & ResidualEncoderUNet \\
Number of stages & 6 \\
Features per stage & [32, 64, 128, 256, 320, 320] \\
Kernel sizes & $3 \times 3 \times 3$ (all stages) \\
Strides & [1,1,1], [2,2,2] (×5) \\
Blocks per stage & [1, 3, 4, 6, 6, 6] \\
Decoder convs & [1, 1, 1, 1, 1] \\
Normalization & InstanceNorm3d  \\
Nonlinearity & Leaky ReLU  \\
Bias & True \\
\hline
\multicolumn{2}{l}{\textbf{Training Configuration}} \\
Batch size & 2 \\
Batch Dice & True \\
\hline
\end{tabular}
\label{tab:nnunet-setup}
\end{table}

\section{Segmentation performance}
\label{app:segmentation}
In this section, we present the segmentation  performance on the three test sets using the Dice similarity coefficient and Hausdorff distance at 95th percentile (Table~\ref{app:seg1}, \ref{app:seg2}, \ref{app:seg3}). Significance tests are only performed for the main metric: Dice similarity coefficient. The statistical analysis plan follows a hierarchical scheme, where only if superiority of the proposed model over TotalSegmentator is established, supriority over the BAMF model is assessed. 

\begin{table}[h]
\centering
\small
\caption{Segmentation results on the TCGA-KIRC test set. PP = post-processing, best values per metric per region are presented in bold. $\dagger$ Significantly superior to TotalSegmentator (p $<$ 0.05).}
\begin{booktabs}{
  colspec = {p{3.5cm}p{2cm}p{2.2cm}},
}
\toprule
  \textbf{Kidney region} & Dice $\uparrow$ & HD95 (mm) $\downarrow$ \\
\midrule
    TotalSegmentator & 0.80 $\pm$ 0.14 & 25.90 $\pm$ 32.41  \\
    Renal-Net (ours) & 0.95 $\pm$ 0.03 & 3.79 $\pm$ 5.89 \\
    Renal-Net + PP (ours) & \textbf{0.95 $\pm$ 0.03}$^{\dagger}$ & \textbf{3.79 $\pm$ 5.89} \\
\midrule
  \textbf{Kidney+mass region} & Dice $\uparrow$ & HD95 (mm) $\downarrow$  \\
\midrule
    TotalSegmentator & 0.75 $\pm$ 0.18 & 28.50 $\pm$ 26.90  \\
    Renal-Net (ours) & 0.97 $\pm$ 0.02 & 3.64 $\pm$ 7.85 \\
    Renal-Net + PP (ours) & \textbf{0.97 $\pm$ 0.02}$^{\dagger}$ & \textbf{3.64 $\pm$ 7.85} \\
\midrule
  \textbf{Renal mass region} & Dice $\uparrow$ & HD95 (mm) $\downarrow$  \\
\midrule
    TotalSegmentator & 0.05 $\pm$ 0.13 & 120.18 $\pm$ 61.30 \\
    Renal-Net (ours) & 0.86 $\pm$ 0.25 & 14.94 $\pm$ 25.81 \\
    Renal-Net + PP (ours) & \textbf{0.86 $\pm$ 0.25}$^{\dagger}$ & \textbf{14.94 $\pm$ 25.81} \\
\bottomrule
\end{booktabs}
\label{app:seg1}
\end{table}

\begin{table}[h!]
\centering
\small
\caption{Segmentation results on the Charité Universitätsmedizin Berlin test set. PP = post-processing, best values per metric per region are presented in bold. $\ddagger$~Significantly superior to BAMF model (p $<$ 0.05.)}
\begin{booktabs}{
  colspec = {p{3.5cm}p{2cm}p{2.2cm}},
}
\toprule
  \textbf{Kidney region} & Dice $\uparrow$ & HD95 (mm) $\downarrow$ \\
\midrule
    BAMF-model & 0.83 $\pm$ 0.13 & 12.19 $\pm$ 18.90 \\
    Renal-Net (ours) & 0.85 $\pm$ 0.10 & 9.84 $\pm$ 16.11 \\
    Renal-Net + PP (ours) & \textbf{0.85 $\pm$ 0.10}$^{\ddagger}$ & \textbf{9.84 $\pm$ 16.11} \\
\midrule
  \textbf{Kidney+mass region} & Dice $\uparrow$ & HD95 (mm) $\downarrow$ \\
\midrule
    BAMF-model & 0.89 $\pm$ 0.11 & 10.43 $\pm$ 13.68 \\
    Renal-Net (ours) & 0.90 $\pm$ 0.07 & 9.28 $\pm$ 12.40 \\
    Renal-Net + PP (ours) & \textbf{0.90 $\pm$ 0.08}$^{\ddagger}$ & \textbf{9.21 $\pm$ 11.92} \\
\midrule
  \textbf{Renal mass region} & Dice $\uparrow$ & HD95 (mm) $\downarrow$  \\
\midrule
    BAMF-model & 0.81 $\pm$ 0.19 & 13.43 $\pm$ 18.47 \\
     Renal-Net (ours) & 0.83 $\pm$ 0.17 & 10.75 $\pm$ 16.86 \\
    Renal-Net + PP (ours) & \textbf{0.83 $\pm$ 0.17}$^{\ddagger}$ & \textbf{10.46 $\pm$ 15.83} \\
\bottomrule
\end{booktabs}
\label{app:seg2}
\end{table}

\begin{table*}[h]
\small
\centering
\caption{Segmentation results on the Radboudumc test set. B20 is a subset comprising of 20 patients with healthy kidneys. B30 is a subset comprising of 30 patients with kidneys containing lesions. PP = post-processing, best values per metric per region are presented in bold. $\dagger$ Significantly superior to TotalSegmentator (p $<$ 0.05). $\ddagger$ Significantly superior to BAMF model (p $<$ 0.05, tested conditional on superiority over TotalSegmentator).}
\begin{tabular}{p{4.5cm} l l l l}
\toprule
  \textbf{Kidney region}
    & \multicolumn{2}{c}{B20}
    & \multicolumn{2}{c}{B30} \\
\midrule
    & Dice $\uparrow$ & HD95 (mm) $\downarrow$
    & Dice $\uparrow$ & HD95 (mm) $\downarrow$ \\
\cmidrule(lr){2-3} \cmidrule(lr){4-5}
  TotalSegmentator   & \textbf{0.95 $\pm$ 0.01} & \textbf{7.39 $\pm$ 23.71} & 0.92 $\pm$ 0.10 & \textbf{15.58 $\pm$ 37.87} \\
  BAMF model    & 0.90 $\pm$ 0.02 & 57.83 $\pm$ 108.77 & 0.90 $\pm$ 0.04 & 18.44 $\pm$ 43.61 \\
   Renal-Net (ours) & 0.93 $\pm$ 0.06 & 55.80 $\pm$ 101.21 & 0.94 $\pm$ 0.05 & 44.32 $\pm$ 99.98  \\
  Renal-Net + PP (ours)  & 0.93 $\pm$ 0.06 & 55.80 $\pm$ 101.21 & \textbf{0.94 $\pm$ 0.05}$^{\dagger\ddagger}$ & 44.32 $\pm$ 99.98  \\
\midrule
  \textbf{Kidney+mass region}
    & \multicolumn{2}{c}{B20}
    & \multicolumn{2}{c}{B30} \\
\midrule
    & Dice $\uparrow$ & HD95 (mm) $\downarrow$
    & Dice $\uparrow$ & HD95 (mm) $\downarrow$ \\
\cmidrule(lr){2-3} \cmidrule(lr){4-5}
  TotalSegmentator     & – & –         & 0.94 $\pm$ 0.02 & \textbf{15.40 $\pm$ 37.83}  \\
  BAMF model           & – & –         & 0.84 $\pm$ 0.15 & 70.58 $\pm$ 88.95  \\
  Renal-Net (ours)       & – & –         & 0.94 $\pm$ 0.05 & 49.44 $\pm$ 102.72  \\
  Renal-Net + PP (ours)  & – & –         & \textbf{0.95 $\pm$ 0.02}$^{\dagger\ddagger}$ & 43.66 $\pm$ 100.00 \\
\midrule
  \textbf{Renal Mass region}
    & \multicolumn{2}{c}{B20}
    & \multicolumn{2}{c}{B30} \\
\midrule
    & Dice $\uparrow$ & HD95 (mm) $\downarrow$
    & Dice $\uparrow$ & HD95 (mm) $\downarrow$ \\
\cmidrule(lr){2-3} \cmidrule(lr){4-5}
  TotalSegmentator     & – & –         & 0.06 $\pm$ 0.14 & 123.27 $\pm$ 113.94 \\
  BAMF model           & – & –         & 0.45 $\pm$ 0.35 & 119.28 $\pm$ 113.74 \\
  Renal-Net (ours)       & – & –         & 0.64 $\pm$ 0.31 & 50.41 $\pm$ 76.18 \\
  Renal-Net + PP (ours)  & – & –         & \textbf{0.66 $\pm$ 0.29}$^{\dagger\ddagger}$ & \textbf{31.84 $\pm$ 39.10}  \\
\bottomrule
\end{tabular}
\label{app:seg3}
\end{table*}

\section{Detection results} \label{app:detection}
Table~\ref{tab:threshold0} presents the detection performance for an overlap threshold of 0 based on the intersection of union. 

\begin{table*}[h]
\small
\centering
\caption{Detection performance with an overlap threshold of 0 based on the intersection over union. PP = post-processing, best values per metric per dataset are presented in bold.}
\begin{booktabs}{
  colspec = {p{4cm}p{2.5cm}p{2.5cm}p{2.5cm}},
}
\toprule
Radboudumc test set & Precision & Recall & F1-score  \\
    \midrule
    TotalSegmentator & 0.42 ($\pm$ 0.40) & 0.11 ($\pm$ 0.26) & 0.23 ($\pm$ 0.21) \\
    BAMF-model & 0.58 ($\pm$ 0.31) & \textbf{0.66 ($\pm$ 0.28)} & 0.57 ($\pm$ 0.23) \\
    Renal-Net (ours) & 0.76 ($\pm$ 0.32) & 0.66 ($\pm$ 0.29) & 0.67 ($\pm$ 0.26) \\
    Renal-Net + PP (ours) & \textbf{0.83 ($\pm$ 0.23)} & 0.64 ($\pm$ 0.32) &  \textbf{0.74 ($\pm$ 0.20)} \\
    \midrule
TCGA-KIRC & Precision & Recall & F1-score  \\
\midrule
    TotalSegmentator & 0.76 ($\pm$ 0.34) & 0.28 ($\pm$ 0.35) & 0.51 ($\pm$ 0.24) \\
    Renal-Net (ours) & 0.91 ($\pm$ 0.18) & 0.69 ($\pm$ 0.33) & 0.79 ($\pm$ 0.22)  \\
    Renal-Net + PP (ours) & \textbf{0.91 ($\pm$ 0.18)} & \textbf{0.69 ($\pm$ 0.33)} & \textbf{0.79 ($\pm$ 0.22)} \\
\midrule 
Charité Universitäts- \newline medizin Berlin & Precision & Recall & F1-score  \\
\midrule
    BAMF-model & 0.78 ($\pm$ 0.27) & 0.91 ($\pm$ 0.23) & 0.81 ($\pm$ 0.23) \\
    Renal-Net (ours) & 0.83 ($\pm$ 0.26) & \textbf{0.92 ($\pm$ 0.22)} & 0.84 ($\pm$ 0.22) \\ 
    Renal-Net + PP (ours) & \textbf{0.84 ($\pm$ 0.25)} & 0.91 ($\pm$ 0.22) & \textbf{0.85 ($\pm$ 0.22)} \\ 
    \bottomrule
\end{booktabs}
\label{tab:threshold0}
\end{table*}

\clearpage

\section{Subgroup analysis} \label{app:subgroup-analysis}
\begin{figure}[h]
    \centering
    \includegraphics[width=0.9\linewidth]{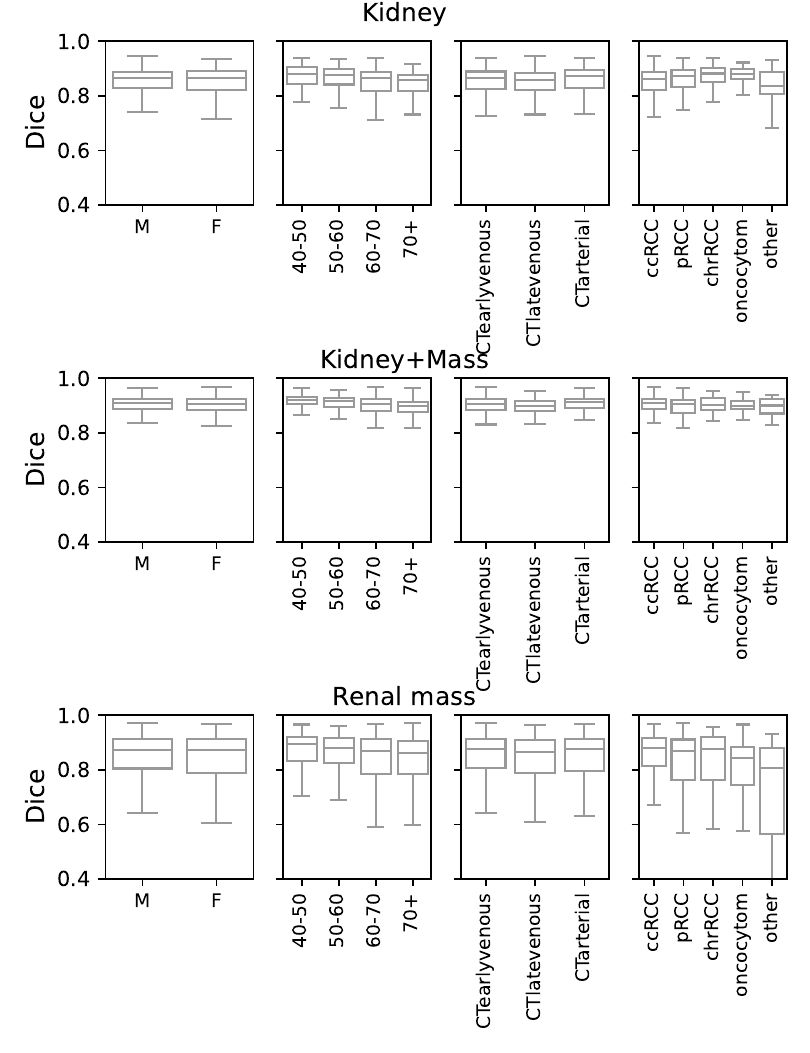}
    \caption{Subgroup analysis of the BAMF-model performed on the Charité Universitätsmedizin Berlin test set. Subgroups that are tested for are patient sex, patient age, CT contrast phase and tumor histologic subtype. We show performance measured in Dice. The boxplots show the median (horizontal line in the box), quartiles (presented by the box) and the full distribution (presented by the whiskers). For the purpose of readability we omitted the outliers from these plots.}
    \label{fig:subgroup-bamf}
\end{figure}

In this section, we present an additional subgroup analysis of the BAMF model on the Charité Universitätsmedizin Berlin test set. Figure~\ref{fig:subgroup-bamf} presents these results and it can be noted that similarly as the proposed model, the BAMF model does not seem to have a systematic bias towards a subgroup. 

\section{Qualitative results} \label{app:qualitative}
\begin{figure*}[h]
    \centering
    \includegraphics[width=0.8\linewidth]{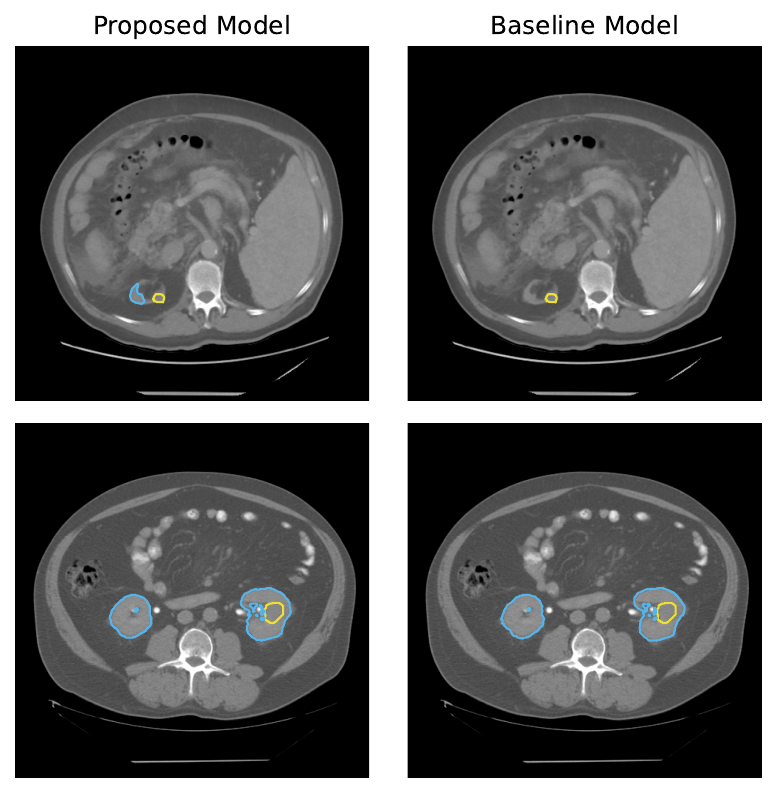}
    \caption{Presented are two cases (top and bottom) from the Charité Universitätsmedizin Berlin test set and the predictions of the proposed model and the BAMF-model (baseline). The reference renal mass mask \textcolor{quali_yellow}{\rule[-0.3mm]{.3cm}{.3cm}} is shown, both models were not able to detect and segment these renal masses. The predicted kidneys \textcolor{quali_blue}{\rule[-0.3mm]{.3cm}{.3cm}} are also highlighted.}
    \label{fig:case1}
\end{figure*}

In this section, we present additional qualitative results for the Charité Universitätsmedizin Berlin test set. In Figure~\ref{fig:case1}, both models were able to correctly delineate the kidneys (blue) but failed to segment the clear cell renal cell carcinomas (yellow ground truth). The tumor in the first row is clearly visible; it is likely the small form of the remnant kidney that confuses both models. The tumor in the second row has minimal contrast differences to the kidney, it is very hard to detect even if the reference annotation mask is known.

\begin{figure*}[h]
    \centering
    \includegraphics[width=0.8\linewidth]{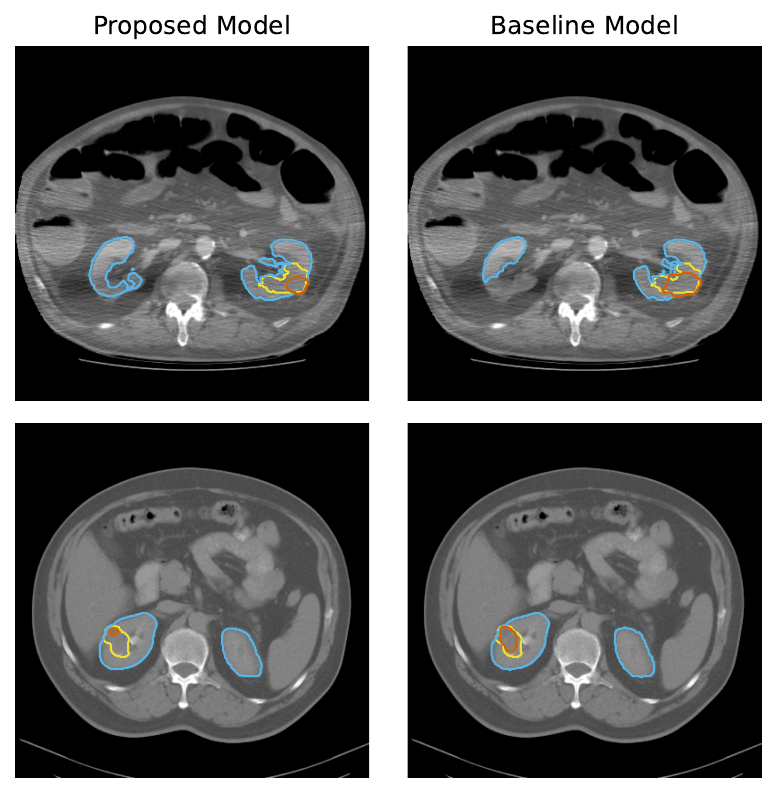}
    \caption{Presented are two cases (top and bottom) from the Charité Universitätsmedizin Berlin test set and the predictions of the proposed model and the BAMF-model (baseline). The reference renal mass mask \textcolor{quali_yellow}{\rule[-0.3mm]{.3cm}{.3cm}} and the renal mass predictions \textcolor{quali_orange}{\rule[-0.3mm]{.3cm}{.3cm}} are shown. The predicted kidney \textcolor{quali_blue}{\rule[-0.3mm]{.3cm}{.3cm}} is also highlighted.}
    \label{fig:case3}
\end{figure*}

In Figure~\ref{fig:case3}, both models correctly segment the kidneys (blue) and are able to localize the tumors (orange). However, the BAMF-model (baseline) was better in capturing the whole tumor. The scan in the upper row has visible horizontal lines through the scan, possibly caused by metal.

\begin{figure*}[h]
    \centering
    \includegraphics[width=0.9\linewidth]{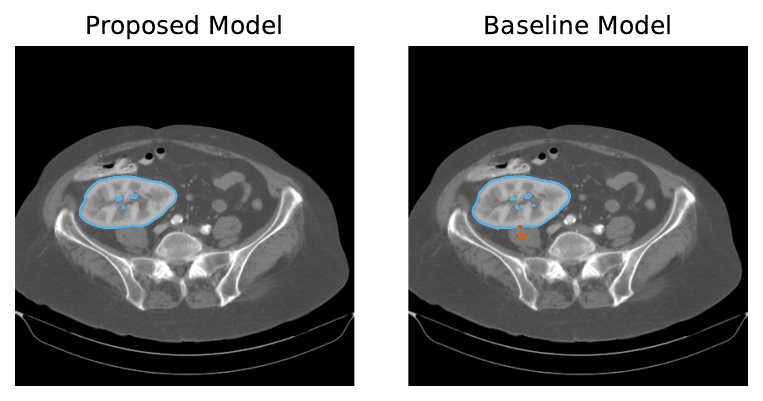}
    \caption{Presented is an interesting transplantation case from the Charité Universitätsmedizin Berlin test set and the predictions of the proposed model and the BAMF-model (baseline). The predicted kidney \textcolor{quali_blue}{\rule[-0.3mm]{.3cm}{.3cm}} is highlighted and shows the robustness of the model, despite the location being in the pelvic area.}
    \label{fig:case4}
\end{figure*}

In Figure~\ref{fig:case4}, we present a case where both models were able to correctly segment a transplanted kidney (blue) despite its uncommon orientation and location in the pelvic area. 

\end{document}